# MKDTI: Predicting drug-target interactions via multiple kernel fusion on graph attention network

Yuhuan Zhou[1], Yulin Wu[1], Weiwei Yuan[1], Xuan Wang[1, 2] and Junyi Li[1, 2, *]

[1]School of Computer Science and Technology, Harbin Institute of Technology (Shenzhen), Shenzhen, Guangdong 518055, China
[2]Guangdong Provincial Key Laboratory of Novel Security Intelligence Technologies, Harbin Institute of Technology (Shenzhen), Shenzhen, Guangdong 518055, China

* To whom correspondence should be addressed. Tel: +86 577 26705201; Email: lijunyi@hit.edu.cn

*Abstract*—Drug-target relationships may now be predicted computationally using bioinformatics data, which is a valuable tool for understanding pharmacological effects, enhancing drug development efficiency, and advancing related research. A number of structure-based, ligand-based and network-based approaches have now emerged. Furthermore, the integration of graph attention networks with intricate drug-target studies is an application area of growing interest. In our work, we formulate a model called MKDTI by extracting kernel information from various layer embeddings of a graph attention network. This combination improves the prediction ability with respect to novel drug-target relationships. We first build a drug-target heterogeneous network using heterogeneous data of drugs and targets, and then use a self-enhanced multi-head graph attention network to extract potential features in each layer. Next, we utilize embeddings of each layer to computationally extract kernel matrices and fuse multiple kernel matrices. Finally, we use a Dual Laplacian Regularized Least Squares framework to forecast novel drug-target entity connections. This prediction can be facilitated by integrating the kernel matrix associated with the drug-target. We measured our model's efficacy using AUPR and AUC. Compared to the benchmark algorithms, our model outperforms them in the prediction outcomes. In addition, we conducted an experiment on kernel selection. The results show that the multi-kernel fusion approach combined with the kernel matrix generated by the graph attention network provides complementary insights into the model. The fusion of this information helps to enhance the accuracy of the predictions.

## I. INTRODUCTION

Finding the ideal target for a medication is a crucial first step in both drug repositioning [1] and the creation of new medications [2]. However, the target-drug relationship is complex, and the relationship between a target and a drug may not be a one-to-one relationship. To understand this task, we first need to gain a deeper understanding of the underlying disease mechanisms, which can be categorized as endogenous and exogenous. Endogenous diseases arise from imbalances in the normal physiological functions and homeostasis within the human body itself. In contrast, exogenous diseases are triggered by external pathogens invading and disrupting normal bodily functions.

For tackling endogenous diseases, omics approaches such as genomics, proteomics, and metabolomics can be powerful tools. By comparing differences in genomes, proteomes, metabolomes between healthy and diseased tissues or biofluids, researchers can pinpoint key proteins and pathways that have become imbalanced. These then become potential targets for drug interventions aimed at restoring equilibrium.

For exogenous conditions, multi-omics analyses enable the identification of proteins and molecules that are specific to the pathogen or are significantly different from the human host. Intervention with these unique pathogenic factors helps to eliminate or inhibit the growth and reproduction of active microorganisms, mitigating their destructive effects on the human body.

However, simply identifying a potential target protein is insufficient. Detailed studies must be undertaken to validate that modulating the target does indeed produce the intended pharmacological actions for treating the disease, without causing unacceptable side effects [3]. This requires an integrated understanding of the target's role within wider biological networks and pathways, as well as confirmation through experimental methods like gene knockdowns or site-directed mutagenesis. Only once a target's dominant role in overall body regulation is established, can it be considered for further drug development [4].

Computational tools and simulations are also increasingly utilized to predict targets of drug candidates based on different approaches. Structure-based methods leverage experimental or homology-modelled 3D structures of target proteins. Computational techniques like molecular docking and reverse pharmacophore mapping can then identify binding targets and key molecular interactions. Ligand-based approaches involve screening large databases and predicting probable targets by comparing the similarity of novel compounds to those of known bioactive ligands [5-8]. Network-based methods map out elaborate interaction networks between compounds, targets, and diseases to discern patterns and associations even when the target structures are unknown.

However, significant challenges remain in accurately predicting drug-target interactions due to their sheer complexity [9-11]. Within the body, targets are not isolated but interact with one another in intricate ways across pathways and tissues. Furthermore, target proteins demonstrate conformational flexibility and shifts in dynamic equilibrium, making computational modeling difficult. Nevertheless, advancing AI and simulation techniques offer new promise in gaining insights, narrowing down the search, and accelerating the drug discovery pipeline.

By modeling tasks as link prediction problems in heterogeneous networks of biological networks, graph neural networks (GNNs), an emerging force in computing, have emerged as a potent tool for molecular interaction prediction [12-18]. By aggregating information from network neighborhoods, GNNs can effectively capture correlational signals and predict new DTIs as missing edges. Various techniques have been explored, including attention mechanisms, metapath-based reasoning, and graph transformers. In order to enable DTI prediction, Wan et al. [19] proposed the NeoDTI model. NeoDTI uses eight independent drugs(targets) related networks to construct a heterogeneous network in which two nodes may have multiple types of edges simultaneously. It then designs specific aggregation functions for each specific edge type to aggregate neighbourhood information and learn node characteristics. Li et al.'s [20] model IMCHGAN learns two levels of representational features for nodes: first using GAT to learn drug(target) features under a specific meta-path, and then integrating these features using the attention mechanism to obtain final node features. Finally an inductive matrix complementation model is used to predict drug-target interactions. Peng et al.'s [21] EEG-DTI constructs a bio-heterogeneous network containing four kinds of nodes and eight kinds of edges, and uses a three-layer GCN to simultaneously aggregate the information of heterogeneous neighbouring nodes to learn drug and target features. The final DTI is predicted using the inner product method.

Multiple kernel learning (MKL) [22] has recently received a lot of attention because, in addition to GNNs being used to predict bioinformatic network connections, it is often used to improve the efficiency of connection prediction in bioinformatic bipartite graphs. Cen et al. [23] proposed a method that can automatically optimize the multi-kernel combinations of diseases and miRNAs for miRNA-disease association prediction. A multi-kernel combination-based clustering technique was presented by Qi et al. [24] that can find groupings in scRNA-seq data directly. For the purpose of predicting novel drug-target interactions, Yan et al. [25] suggested integrating heterogeneous data pertaining to a variety of drugs and targets. A multi-kernel fusion graph convolutional network (MKGCN) model was presented by Yang et al. [26] to infer novel microbe-drug interactions.

Multiple kernel learning (MKL) is a powerful technique for combining diverse information to improve predictive modeling. By integrating multiple kernel matrices, each capturing different aspects of the data, MKL can boost model performance. However, the success of MKL depends on constructing informative kernel matrices from high-quality, representative features. This is challenging for drug-target prediction.

Recent graph neural networks (GCN [27], GAT [28]) offer a solution through multi-layer network embeddings. As neighborhood information propagates through GNN layers, the learned node embeddings encode successively broader network contexts. The embeddings at each layer thus represent different semantic information. This provides a natural way to generate multiple diverse kernel matrices for MKL, by extracting similarity metrics between node embeddings across layers.

In conclusion, we combine a multi-kernel fusion method with graph attention networks to present an end-to-end prediction model for drug-target relationships. First, we combine the drug-target association bipartite graph with the similarity matrix between drugs and the similarity matrix between targets to create a heterogeneous network. We randomly initialize the node feature matrices. Then we extract the node embeddings of different layers by graph attention network layers and extract the kernel matrix corresponding to this embedding. Finally, a Dual Laplacian Regularized Least Squares(DLapRLS) [29] model uses the final drug and target embedding vectors that we get via a multi-kernel fusion technique to forecast unknown connections between drugs and targets. We downloaded the recent drug and target heterogeneity data to compare MKDTI with the latest method. Our model outperforms other baselines on both AUC and AUPR, demonstrating good predictive power. In addition, we conducted an experimental study of kernel selection, and the results showed that the multi-kernel fusion model can be combined with graph-attention networks to generate kernel matrices, which provide complementary information to the model and improve the prediction results.

II. DATASETS AND MATERIALS

A. Dataset

Finding new treatments requires being able to anticipate drug-target interactions accurately, but one of the biggest obstacles is the paucity of comprehensive information on the vast biological drug-target space. Computational approaches trained on incomplete datasets struggle to reliably generalize to undiscovered interactions.

Many new drugs and targets have been discovered in recent years, as well as new experimentally demonstrated drug-target associations. We construct a more comprehensive training set for computational DTI prediction by using data derived from the recent releases of major public databases. Specifically, we collected a wide range of unique drug information, drug-drug interaction, drug-target associations

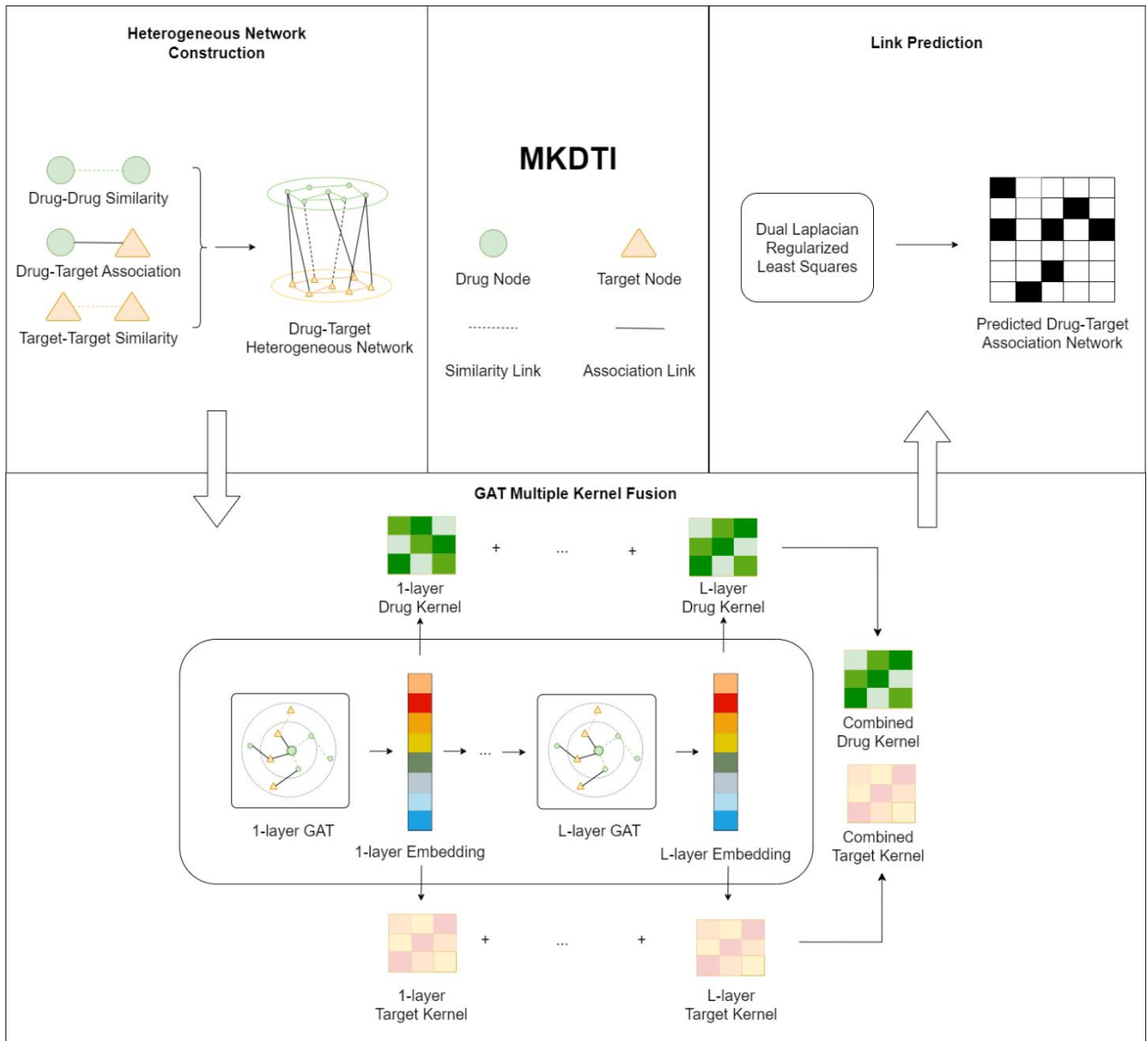

Fig.1 Flowchar. Heterogeneous network construction module - the drug-target heterogeneous network is collaboratively built utilizing drug similarity, target similarity, and drug-target associations; GAT multi-kernel fusion module - using input vectors for multi-layer GAT training, for each layer of the learned embeddings to extract kernel features, and then use average weights to fuse multiple kernel matrices; Link Prediction Module - applies the DLapRLS model to predict association.

from DrugBank [30]. In addition, we collected Target-Target interactions from STRING [31]. Finally we collected human target information from UniProtKB [32]. In the end, The statistics for our use of the dataset are as follows:

TABLE I. DATASETS STATISTICS

| Items | Numbers | Resources |
|---|---|---|
| **Drug** | 1956 | DrugBank |
| **Target** | 1726 | UniProtKB |
| Drug-Drug | 939585 | DrugBank |
| Target-Target | 344110 | STRING |
| Drug-Target | 6679 | DrugBank |

### B. Evaluation Indicators

AUPR and AUC are often used metrics to assess how well a link prediction task is performing. By characterizing the area under the curve of the true-positive rate vs the false-positive rate, AUC quantifies the capacity to discriminate between positives and negatives. Through the characterization of the area under the precision vs. recall curve, AUPR places more emphasis on positive class performance. Combining both metrics provides fuller assessment of model quality on imbalanced link prediction problems.

$$TruePositiveRate = \frac{TP}{TP + FN} \quad (1)$$

$$FalsePositiveRate = \frac{FP}{FP + TN} \quad (2)$$

## III. MODEL AND IMPLEMENTATION

### A. Problem Description

Drug-target association prediction aims to predict whether there is an interaction between a pair of a drug molecule and a target molecule. We modelled the task as a heterogeneous network complementation task: drug node A and target node B are considered to interact if there is an edge between them in the network predicted by the model.

Specifically, the heterogeneous network we construct defines two types of nodes: drug nodes $d_i (1 \leq i \leq N_d)$ and target nodes $t_j (1 \leq j \leq N_t)$. $N_d$ and $N_t$ represent the number of drug nodes and target nodes in the network respectively. The set of drug nodes of this heterogeneous network is denoted as D:

$$D = \{d_i \mid 1 \leq i \leq N_d\} \quad (3)$$

The set of target nodes of this heterogeneous network is denoted as T:

$$T = \{t_j \mid 1 \leq j \leq N_t\} \quad (4)$$

The associations of drugs and targets in a heterogeneous network is denoted as a heterogeneous network incidence matrix $Y \in R^{N_d \times N_t}$. We use matrix element values $Y_{i,j}$ equal to 1 to indicate the existence of an edge between two nodes, and matrix element values $Y_{i,j}$ equal to 0 to indicate that the edge between two nodes is unknown. We can predict unknown drug-target associations by outputting a new association network $Y^* \in R^{N_d \times N_t}$.

### B. Drug-Target Heterogeneous Network

The Tanimoto score [33] was derived from the drug SMILES string in order to create the drug similarity. By computing the Jaccard similarity coefficient in the target-target interaction matrix, target similarity is created. $K_s^d$ represents the drug similarity between drugs $\{d_i, d_j\} \in D$. $K_s^t$ represents the target similarity between targets $\{t_i, t_j\} \in T$.

Next, we constructed Y, $K_s^t$, and $K_s^d$ as the three components of a heterogeneous biological network.

We use the adjacency matrix $A \in R^{(N_d+N_t) \times (N_d+N_t)}$ to represent the heterogeneous networks:

$$A = \begin{bmatrix} K_s^d & Y \\ Y^T & K_s^t \end{bmatrix} \quad (5)$$

### C. Self-Augmented Multi-Headed Graph Attention Networks

Graph Attention Network(GAT) [28] introduce self-attention and multi-attention techniques in contrast to Graph Convolutional Networks(GCN) [27] in order to better capture dependencies between nodes.

In GAT, each node updates its own representation by interacting with its neighboring nodes and aggregates the information from different neighboring nodes by attaching weights to them based on the computed attention scores. In GAT, each node can have multiple attention heads, and each header is able to learn different node weights so as to adaptively adjust the importance of the node during the information transfer process. We calculate the attention coefficient as follows:

$$e_{ij} = a(W\vec{h}_i, W\vec{h}_j) \quad (6)$$

W is a learnable weight matrix that can be linearly transformed with respect to the node features h. $a(\cdot)$ is a function that calculates the correlation between two nodes (feature vectors). Next normalize using softmax and add LeakyReLU to provide nonlinearity.

$$\alpha_{i,j} = softmax_j(e_{ij}) = \frac{exp(e_{ij})}{\sum_{k \in N_i} exp(e_{ik})} \quad (7)$$

$$\alpha_{i,j} = \frac{exp(LeakyReLU(\vec{a}^T[W\vec{h}_i || W\vec{h}_j]))}{\sum_{k \in N_i} exp(LeakyReLU(\vec{a}^T[W\vec{h}_i || W\vec{h}_k]))} \quad (8)$$

Using the self-enhancing attention mechanism, the current node is treated as its own neighbor to obtain a single-head attention node embedding for each layer:

$$\vec{h}_i' = \sigma(\sum_{j \in N_i} \alpha_{ij} W\vec{h}_j) \quad (9)$$

Next, node features are computed using k independent attention mechanisms and then aggregated by concatenation or averaging so that the information contained in the features can be extended:

$$\vec{h}_i' = \mathop{\|}_{k=1}^{K} \sigma(\sum_{j \in N_i} \alpha_{ij}^k W^k \vec{h}_j) \quad (10)$$

Or

$$\vec{h}_i' = \sigma(\frac{1}{K} \sum_{k=1}^{K} \sum_{j \in N_i} \alpha_{ij}^k W^k \vec{h}_j) \quad (11)$$

Here, $\alpha^k$ means the normalized attention coefficient of the k th attention head, $\sigma$ means the normalization function sigmoid, $\|$ means the concatenation operation.

### D. Multi-Kernel Fusion based on Graph Attention Network

Multilayer GAT models can compute multiple embeddings that represent information about different graph structures. During each layer of information transfer, the original node features $H_0$ will continuously aggregate the neighbor information based on the attention weights. After multilayer network transmission, L-order node feature $H_l$ includes their L-order neighbor information. Because each layer of embedded information has a different meaning, we use each layer of GAT embedding as the input vector for calculating the kernel matrices.

We take $N_d$ rows of the embedding vectors of graph attention network l-layer as drug embedding vectors $H_l^d$ and $N_t$ rows as target embedding vectors $H_l^t$, and use Gaussian interaction profile to obtain the kernel matrix of the corresponding embedding vectors. Set the corresponding bandwidth parameter $\gamma_{h_l}$.

$$K_{h_l}^d = \exp(-\gamma_{h_l}\|H_l^d(i) - H_l^d(j)\|^2) \quad (12)$$

$$K_{h_l}^t = \exp(-\gamma_{h_l}\|H_l^t(i) - H_l^t(j)\|^2) \quad (13)$$

By collecting the extracted kernel matrices from different layers, we have a collection of kernel matrices for drugs and targets.

$$S^d = \{K_s^d, K_{h_1}^d, \ldots, K_{h_L}^d\} \quad (14)$$

$$S^t = \{K_s^t, K_{h_1}^t, \ldots, K_{h_L}^t\} \quad (15)$$

As we mentioned before, these kernel matrices contain semantic information about various hop-neighbor aggregations, thereby enhancing the model's predictive capability. We use weighting coefficient to combine different kernels, with the following equation:

$$K_d = \sum_{i=1}^{L+1} \omega_i^d S_i^d \quad (16)$$

Similarly, the target fusion kernel can be computed by the following equation

$$K_t = \sum_{i=1}^{L+1} \omega_i^t S_i^t \quad (17)$$

The $S_i^d$ and $S_i^t$ are matrices of the ith kernels in the collection of kernel matrices for drugs and targets, and $\omega_i^d$ and $\omega_i^t$ are the weights corresponding to each kernel matrices, which in this study is set $\omega_i^d = \omega_i^t = \frac{1}{L+1}$.

*E. Link Prediction Module and Loss Function*

To further enhance the prediction performance of DTIs, Ding et al. [29] suggest a Dual Laplacian Regularized Least Squares (DLapRLS) framework, which is inspired by LapRLS [34]. We use DLapRLS framework to predict drug-target interactions.

Firstly, we compute the normalised Laplace matrices $L_d \in R^{N_d \times N_d}$ and $L_t \in R^{N_t \times N_t}$ by:

$$L_d = D_d^{-1/2} \Delta_d D_d^{1/2}, \Delta_d = D_d - K_d \quad (18)$$

$$L_t = D_t^{-1/2} \Delta_t D_t^{1/2}, \Delta_t = D_t - K_t \quad (19)$$

Here, $K_d \in R^{N_d \times N_d}$ and $K_t \in R^{N_t \times N_t}$ are the fusion kernels. $D_d(k,k)$ and $D_t(k,k)$ are diagonal matrices and are computed by:

$$D_d(k,k) = \sum_{m=1}^{N_d} K_d(k,m) \quad (20)$$

$$D_t(k,k) = \sum_{m=1}^{N_t} K_t(k,m) \quad (21)$$

Let $\|\cdot\|_F$ be the F-paradigm, $Y_{train} \in R^{N_d \times N_t}$ be the association matrix in the training set; $\alpha_d \in R^{N_d \times N_t}$ and $\alpha_t^T \in R^{N_d \times N_t}$ be the trainable matrices. The loss function is shown below:

$$\min J = \|K_d\alpha_d + (K_t\alpha_t)^T - 2Y_{train}\|_F^2 + \lambda_d tr(\alpha_d^T L_d \alpha_d) + \lambda_t tr(\alpha_t^T L_t \alpha_t) \quad (22)$$

Using the trainable matrices $\alpha_d$ and $\alpha_t$ combined with the information from the fusion kernel, the drug-target association prediction results Y* are as follows:

$$Y^* = \frac{K_d\alpha_d + (K_t\alpha_t)^T}{2} \quad (23)$$

During the forward propagation stage, we first establish the model with all trainable parameters and then compute the model's loss function. Using the Adam optimizer [35] to optimize the graphical attention network's parameters and the partial derivatives of the loss function as the DLapRLS framework to iteratively optimize the function are the two types of optimization techniques we employ during back propagation to optimize the loss function.

The following is the formula for the DLapRLS loss function's partial derivative with regard to parameter $\alpha_d$:

$$\frac{\partial J}{\alpha_d} = 2K_d(K_d\alpha_d + \alpha_t^T K_t^T - 2Y_{train}) + 2\lambda_d L_d \alpha_d \quad (24)$$

Let $\frac{\partial J}{\alpha_d}$ be equal to 0:

$$\alpha_d = (K_d K_d + \lambda_d L_d)^{-1} K_d [2Y_{train} - \alpha_t^T K_t^T] \quad (25)$$

Similarly, the following is the formula for the DLapRLS loss function's partial derivative with regard to parameter $\alpha_t$:

$$\frac{\partial J}{\alpha_t} = 2K_t(K_t\alpha_t + \alpha_d^T K_d^T - 2Y_{train}) + 2\lambda_t L_t \alpha_t \quad (26)$$

Let $\frac{\partial J}{\alpha_t}$ be equal to 0:

$$\alpha_t = (K_t K_t + \lambda_t L_t)^{-1} K_t [2Y_{train} - \alpha_d^T K_d^T] \quad (27)$$

IV. EXPERIMENT

*A. Parameter Setting*

In our study, we adjusted the hyperparameters according to the model performance. Consequently, we determine that the model's graph attention network has L = 3 layers, 8 attention heads per layer, and 384,256,128 embedding output dimensions for each layer. Using 5-fold cross-validation and two evaluation indicators, AUPR and AUC, we run 20 iterations of the model with a learning rate of 0.001.

In addition, for the parameters $\lambda_d$、$\lambda_t$、$\gamma_{h_l}(l = 1, \ldots, L)$, we refer to the related literature and conduct experiments, and finally choose the parameters as $\gamma_{h_1} = 2^{-5}, \gamma_{h_2} = 2^{-3}, \gamma_{h_3} = 2^{-3}$. And set $\lambda_d = 2^{-3}, \lambda_t = 2^{-4}$.

*B. Baseline Comparison*

We used our dataset to compare MKDTI with several other advanced drug-target association prediction techniques to confirm the efficacy of our method. These methods include:

NeoDTI [19]: NeoDTI integrates different heterogeneous information from multiple individual networks and defines a neighborhood information aggregation operator to learn node features.

HampDTI [36]: Using a learnable metapath attention mechanism to avoid the drawbacks of manually defining metapaths. Learning multi-channel embeddings and fusion using graph neural networks on extracted multiple metapath graphs.

IMCHGAN [20]: IMCHGAN learns two levels of representational features for nodes: first using GAT to learn drug-target features under a specific meta-path, and then integrating these features using the attention mechanism to obtain final node features. In order to predict drug-target interactions, an inductive matrix complementation model is finally employed.

EEG-DTI [21]: EEG-DTI builds a bio-heterogeneous network with 4 types of nodes and 8 types of edges. It then employs a three-layer GCN to simultaneously collect the data of heterogeneous neighboring nodes. The inner product approach is used to anticipate the final DTI.

SGCL-DTI [37]: In SGCL-DTI, node embeddings are learned and fused based on GCN and attention mechanisms, and drug-target networks are retrieved from heterogeneous networks utilizing metapaths. Drugs and targets are combined into DPP nodes through connections, and topology and semantic graphs are constructed. Finally supervised comparative learning is performed.

With 0.8 percentage point increase in AUC and 1.3 percentage point improvement in AUPR over the second-best performing method, MKDTI exhibits the best performance on the dataset, according to a comparison of the experimental findings displayed in Fig. 2, Tab. 2. This experimental result shows that our heterogeneous network embedding model based on GAT multi-kernel fusion is effective for DTI prediction.

TABLE II. PERFORMANCE COMPARISON WITH OTHER BASELINES

| Methods | AUC | AUPR |
|---|---|---|
| NeoDTI | 0.921 | 0.839 |
| HampDTI | 0.899 | 0.904 |
| IMCHGAN | 0.917 | 0.748 |
| EEG-DTI | 0.928 | 0.949 |
| SGCL-DTI | 0.934 | 0.951 |
| **MKDTI** | **0.942** | **0.964** |

C. Parameters Evaluation

There are several hyperparameters $\gamma_{h_1}, \gamma_{h_2}, \gamma_{h_3}$ and embedding vector dimensions $D_1$、$D_2$、$D_3$ in our model, which have important effects on the performance of our model. $\gamma_{h_1}, \gamma_{h_2}, \gamma_{h_3}$ denote the bandwidth parameters of different layers of the kernel matrix, and different $\gamma_{h_1}, \gamma_{h_2}, \gamma_{h_3}$ lead to different $K_{h_l}^d$ and $K_{h_l}^t$, which affect the effect of the kernel matrix. As we know, the higher the embedding vector dimension, the richer the semantic information that the word vector can encode, and the finer the representation of the word meaning. In practical applications, the embedding vector dimension is too low to easily lose information, and too high to easily introduce noise and increase computational cost.

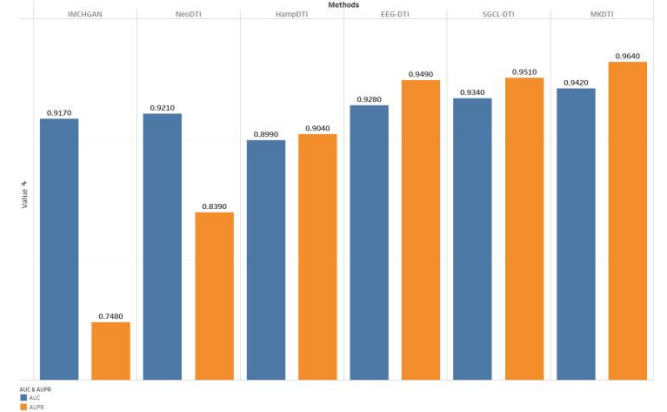

Fig.2 Comparing with the latest baseline on our dataset, MKDTI's AUPR and AUC outperform other models.

To assess the impact of various parameters on the model's performance, we thus do a 5-CV experiment. The performance of the model with various settings is displayed in Fig. 3 and Tab. 3. As can be seen from Fig.3 , when the parameter is larger than a certain threshold, the model performance tends to decrease as the parameter increases. The model's performance decreases significantly when parameter $\gamma_{h_1}$ is increased, but it also somewhat decreases when parameters $\gamma_{h_2}$ and $\gamma_{h_3}$ are increased. After a certain threshold, the model performance also decreases slightly with decreasing parameters. Compared to parameter $\gamma_{h_1}$, parameters $\gamma_{h_2}$ and $\gamma_{h_3}$ have less effect on the model. Furthermore, parameters outside of the range of values will not be considered because when $\gamma_{h_l}$ is too small the kernel matrix will be near to 1. Tab.3 shows the AUPR, AUC values of the model for different $D_i$. It can be seen that the results are worse when the overall is larger.

Finally, based on the experiments, we set the parameter $\gamma_{h_1} = 2^{-5}$、$\gamma_{h_2} = 2^{-3}$、$\gamma_{h_3} = 2^{-3}$ and $D_1 = 384$、$D_2 = 192$、$D_3 = 96$.

TABLE III. PERFORMANCE OF THE MODEL WITH DIFFERENT EMBEDDING DIMENSIONS

| Methods | $D_1$ | $D_2$ | $D_3$ | AUPR | AUC |
|---|---|---|---|---|---|
| MKDTI-1 | 1024 | 512 | 256 | 0.961 | 0.939 |
| MKDTI-2 | 512 | 256 | 128 | 0.963 | 0.942 |
| **MKDTI-3** | **384** | **192** | **96** | **0.964** | **0.942** |
| MKDTI-4 | 256 | 128 | 64 | 0.963 | 0.941 |
| MKDTI-5 | 128 | 64 | 32 | 0.963 | 0.940 |

D. Kernel Matrix Selection Experiments

We will examine the contribution of various kernel matrices to the model's prediction performance and the efficacy of the kernel fusion strategy in this experiment, given that the data from various graph attention network layers serves as a crucial foundation for the creation of kernel matrices.

We used a three-layer GAT in our study, denoted by hl+MKDTI for the kernel matrix obtained by the MKDTI

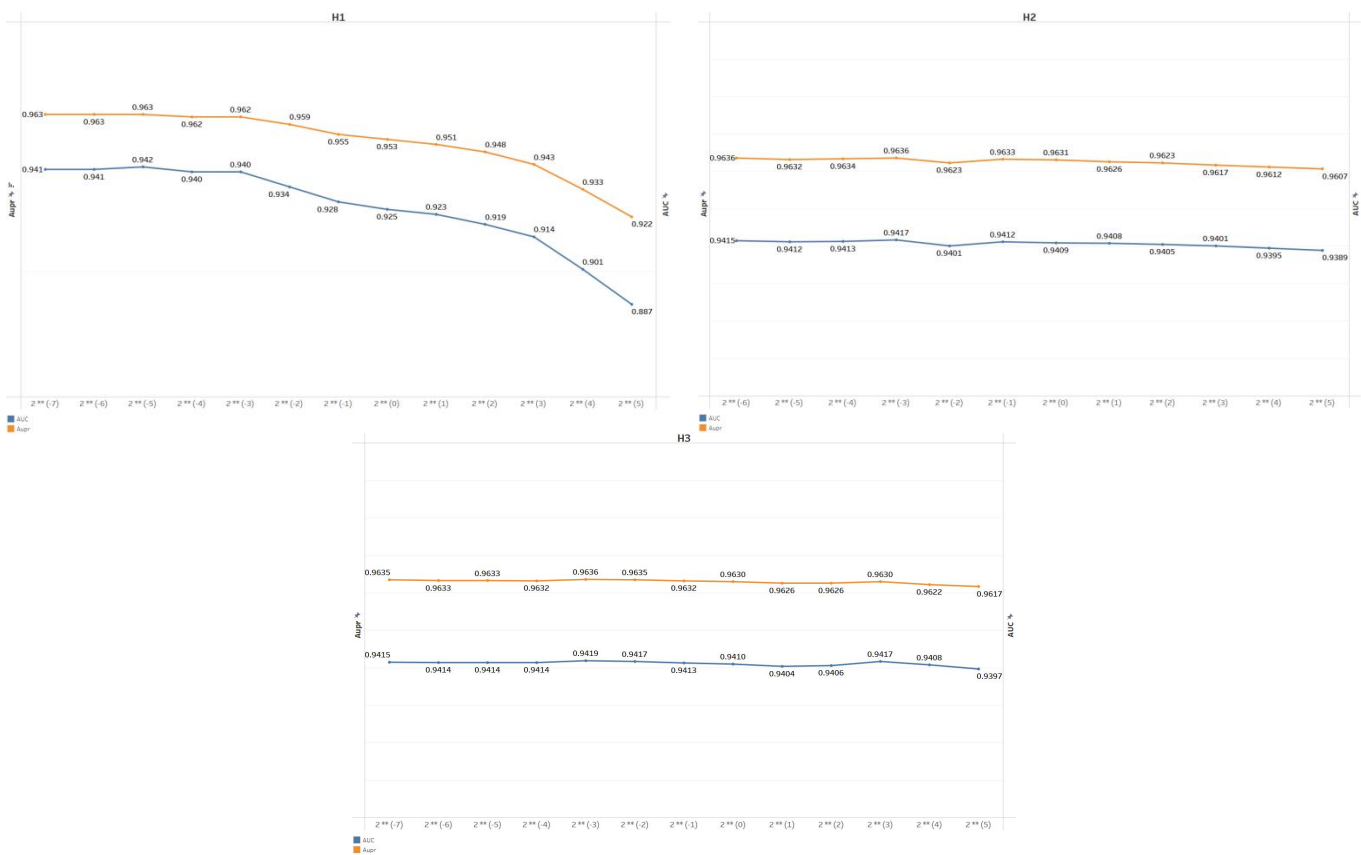

Fig.3 It shows AUPR and AUC of the model under different $\gamma_{h_l}(l = 1, ... , L)$ values. H1 represents $\gamma_{h_1}$, H2 represents $\gamma_{h_2}$, H3 represents $\gamma_{h_3}$.

model using in layer l (l = 1, 2, 3). In addition, we assign weights to the above kernel matrices to construct the MKDTI model.

The AUC, AUPR, and F1 of different models are displayed in Fig.4 and Tab.4:

TABLE IV. PERFORMANCE OF THE MODEL UNDER DIFFERENT KERNEL

| Methods | AUC | AUPR | F1 |
|---|---|---|---|
| h1+MKDTI | 0.936 | 0.947 | 0.873 |
| h2+MKDTI | 0.903 | 0.918 | 0.840 |
| h3+MKDTI | 0.902 | 0.915 | 0.834 |
| **MKDTI** | **0.942** | **0.964** | **0.923** |

Comparison between the single-kernel models reveals that the AUC and AUPR of h1 + MKDTI are significantly higher than those of h2 + MKDTI, and the AUC and AUPR of h2 + MKDTI are slightly higher than those of h3 + MKDTI. i.e., the lower the level of the GAT-generated matrix tends to contain more information, and therefore also achieves better performance.

Comparison between the single-kernel models and MKDTI, MKDTI comprehensively outperforms single-kernel models on AUC, AUPR, and F1. This indicates that the information embedded in the single kernel matrices extracted from different layers of the graph attention network can complement each other. This information can be effectively utilized to enhance the model performance through a multi-kernel fusion approach.

V. CONCLUSION AND DISCUSSION

Target identification is very important for the treatment of diseases, and depending on whether the disease is endogenous or exogenous, different methods can be used to identify key targets. Structure-based, ligand-based, and network-based methods, among others, are all widely used for drug-target association prediction. These techniques assist with drug design, improve drug development success rates, lower R&D expenses, and shorten the R&D cycle; in a similar vein, they aid in comprehending how drugs function at the molecular level. Among them, graph neural network-based methods have received increasing attention and have been used to obtain high-level features in bioinformatic heterogeneous networks. However, existing methods have some problems in data integration and node information aggregation, which limit the prediction performance.

Multi-kernel learning has been proposed recently by researchers to enhance link prediction performance in biobinary networks. The multi-kernel learning method obtains different kernel matrices by diverse information computation and improves the prediction ability by combining information from different kernels. However, obtaining multiple, more representative types of features is a major challenge for kernel matrix applications. In this study, we offer a multi-kernel fusion based link prediction approach for drug-target. The method uses graph attention networks to learn the node embedding features of different layers using different weights to aggregate neighbor node information, and extracts kernel matrices from them to satisfy the requirement of diverse and high-quality information sources for multi-kernel fusion methods. The final drug and target embedding vectors are obtained by the multi-kernel fusion strategy, and these features are utilized for drug-target association prediction. The technique performs exceptionally well in prediction, according to the experimental data, and kernel selection experiment confirm the multi-kernel fusion model's validity.

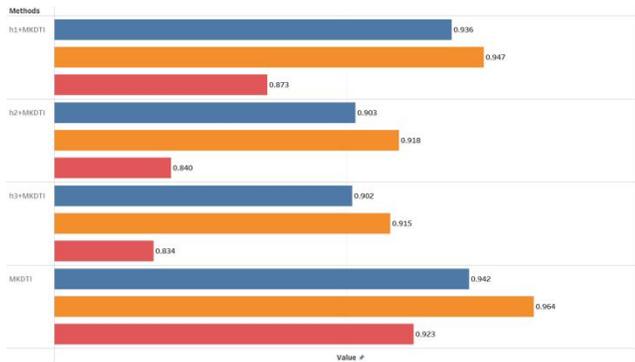

Fig.4 Performance of the model under different kernel

**Competing interests.**
The authors declare that they have no competing interests.

**Additional Files.**
All additional files are available at:
https://github.com/jackedv/MKDTI

**Authors' contributions.**
YZ designed the study, performed bioinformatics analysis and drafted the manuscript. All of the authors performed the analysis and participated in the revision of the manuscript. JL conceived of the study, participated in its design and coordination and drafted the manuscript. All authors read and approved the final manuscript.

VI. ACKNOWLEDGEMENTS.

This work was supported by the grants from the National Key R&D Program of China (2021YFA0910700), Shenzhen Science and Technology Program (JCYJ20200109113201726), Guangdong Basic and Applied Basic Research Foundation (2021A1515012461 and 2021A1515220115), Guangdong Provincial Key Laboratory of Novel Security Intelligence Technologies (2022B1212010005).

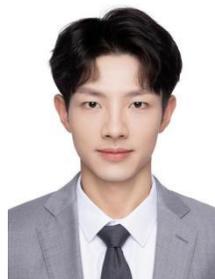

**Yuhuan Zhou** received B.S. degree from Northeastern University and he is a M.S. candidate at Harbin Institute of Technology. His research interests include graph representation learning and biological big data mining.

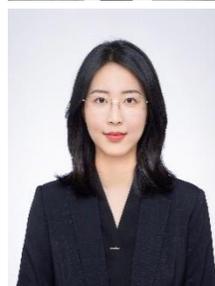

**Yulin Wu** received the Ph.D. degree from Harbin Institute of Technology in

2021. She is currently an assistant professor in Harbin Institute of Technology (Shenzhen), China. Her research interests include artificial intelligence security and applied cryptography.

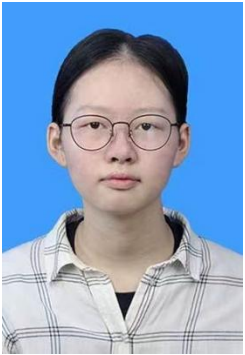

**Weiwei Yuan** has been a bachelor degree candidate in computer science at Harbin Institute of Technology, Shenzhen, China and her research interests include bioinformatic and computer vision.

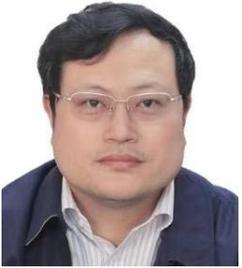

**Xuan Wang** received his M.S. and Ph.D. degrees in Computer Sciences from Harbin Institute of Technology in 1994 and 1997 respectively. He is a professor and Ph.D. supervisor in the Computer Application Research Center, Harbin Institute of Technology, Shenzhen, China. His main research interests include information security, artificial intelligence, and computational linguistics.

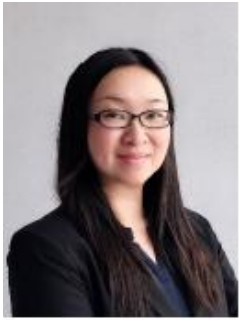

**Junyi Li** received her B.S. degree from Peking University, China and her Ph.D. degree from Rutgers University, the state university of New Jersey, United States. She is an associate professor in the school of computer science and technology at Harbin Institute of Technology (Shenzhen). Her research interests include bioinformatics and system biology.